\documentclass[letterpaper, 10 pt, conference]{ieeeconf}  

\usepackage{algorithm}
\usepackage{algpseudocode}
\usepackage{subcaption}
\usepackage{newfloat}
\usepackage{adjustbox}
\usepackage{listings}
\usepackage{xcolor}
\usepackage{tikz}
\usepackage{mathtools} 
\usepackage{amsfonts} %
\usepackage{cleveref}
\usepackage{svg}
\usepackage{bbm}
\usepackage{pgfplots}
\pgfplotsset{compat=1.18}
\usetikzlibrary[pgfplots.groupplots]

\usepackage{wrapfig}

\usetikzlibrary{arrows, automata, positioning}
\tikzset{auto, >=stealth}
\tikzset{every edge/.append style={shorten >= 1pt}}
\tikzset{
    main node/.style={circle,draw,minimum size=1cm,inner sep=0pt},
}

\DeclareCaptionStyle{ruled}{labelfont=normalfont,labelsep=colon,strut=off} 
\floatstyle{ruled}
\newfloat{listing}{tb}{lst}{}
\floatname{listing}{Listing}

\usepackage{macros}
\usepackage{graphicx, array, blindtext}
\usepackage{dblfloatfix}
\newtheorem{definition}{Definition}

\IEEEoverridecommandlockouts                              
\title{\LARGE \bf
Uncertainty-Guided Enhancement on Driving Perception System via Foundation Models
}

\author{Yunhao Yang$^{1}$, Yuxin Hu$^{2}$, Mao Ye$^{2}$, Zaiwei Zhang$^{2}$, Zhichao Lu$^{2}$, Yi Xu$^{3}$, Ufuk Topcu$^{1}$, Ben Snyder$^{2}$
\thanks{$^{1}$
University of Texas at Austin, Austin, TX, United States 
}
\thanks{$^{2}$
Cruise, San Francisco, CA, United States 
}
\thanks{$^{3}$
Northeastern University, Boston, MA, United States 
}
}

\begin{document}

\maketitle
\thispagestyle{empty}
\pagestyle{empty}

\begin{abstract}
Multimodal foundation models offer promising advancements for enhancing driving perception systems, but their high computational and financial costs pose challenges. We develop a method that leverages foundation models to refine predictions from existing driving perception models---such as enhancing object classification accuracy---while minimizing the frequency of using these resource-intensive models. The method quantitatively characterizes uncertainties in the perception model's predictions and engages the foundation model only when these uncertainties exceed a pre-specified threshold. Specifically, it characterizes uncertainty by calibrating the perception model’s confidence scores into theoretical lower bounds on the probability of correct predictions using conformal prediction. Then, it sends images to the foundation model and queries for refining the predictions only if the theoretical bound of the perception model's outcome is below the threshold. Additionally, we propose a temporal inference mechanism that enhances prediction accuracy by integrating historical predictions, leading to tighter theoretical bounds. The method demonstrates a 10 to 15 percent improvement in prediction accuracy and reduces the number of queries to the foundation model by 50 percent, based on quantitative evaluations from driving datasets.
\end{abstract}
\glsresetall

\section{Introduction}

\begin{figure}
    \centering
    \includegraphics[width=\linewidth]{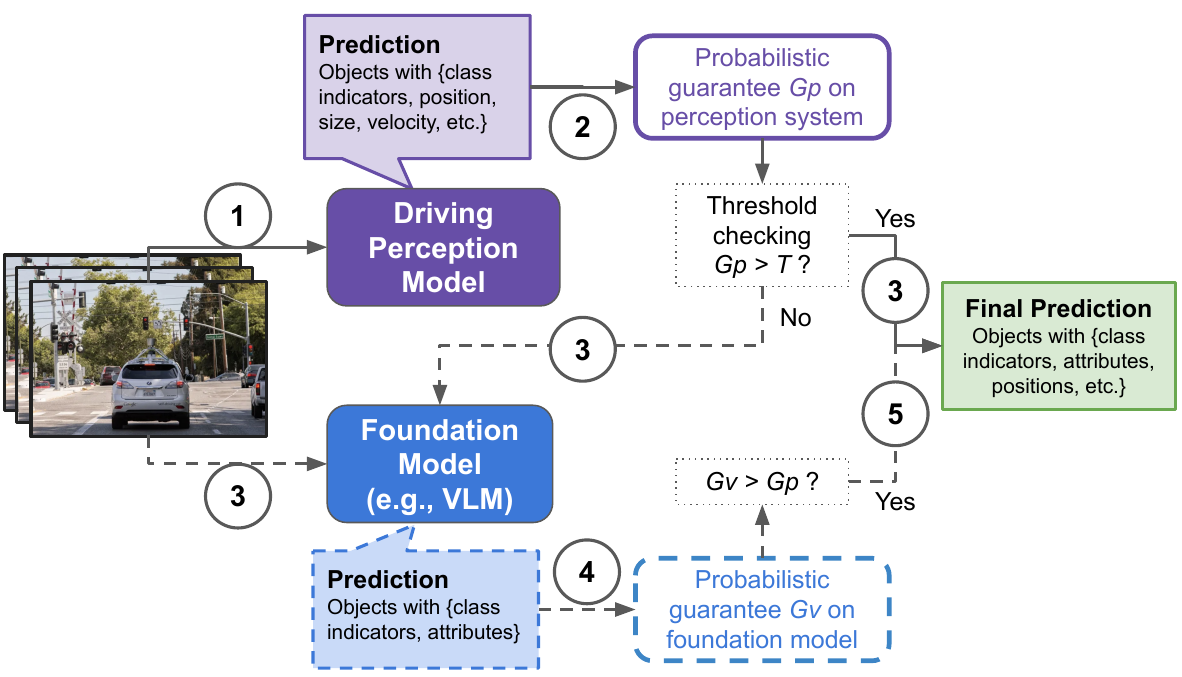}
    \caption{Pipeline of the uncertainty-guided perception system enhancement: \textbf{(1)} The driving perception model takes image-based observations and returns predictions of the observed objects with confidence scores. The predictions include categories, attributes (move or stop), and tracking information (same object across multiple frames). \textbf{(2)} Calibrate each confidence score into a probabilistic guarantee $G_p$. \textbf{(3)} If a probabilistic guarantee $G_p$ is lower than a user-specified threshold $T$, trigger a foundation model to obtain another object prediction. Otherwise, use the perception model's prediction as the final prediction for downstream tasks. \textbf{(4)} Calibrate the foundation model's confidences into probabilistic guarantees $G_v$. \textbf{(5)} If $G_v > G_p$, refine the perception model's prediction with the foundation model's prediction.}
    \label{fig: pipeline}
    \vspace{-15pt}
\end{figure}

In autonomous driving, a driving perception system classifies surrounding objects and predicts their future motions. The ability to accurately classify and predict surrounding objects is crucial for ensuring driving safety. Recent advancements in multimodal foundation models have shown remarkable promise in enhancing the performance of driving perception systems. These models excel in interpreting perception data, such as images, offering improvements in object classification and motion prediction accuracies. However, their widespread adoption faces significant hurdles due to their high computational demands and substantial financial costs, which can limit their practical use in real-time driving scenarios.

Addressing these limitations, we propose a novel method that leverages the strengths of multimodal foundation models while mitigating their operational costs. The method focuses on refining the predictions of existing driving perception models, e.g., improving object classification accuracy, without the need for constant reliance on resource-intensive foundation models. The method utilizes foundation models selectively, based on the uncertainty of the predictions made by the perception system.

Central to the proposed method is the quantitative characterization of uncertainties in the perception model's outputs. The method employs conformal prediction techniques to calibrate the model's confidence scores into theoretical lower bounds on the probability of correct predictions. This bound allows users to assess the reliability of the perception model's predictions rigorously. The method then queries the foundation model only when these theoretical bounds fall below a pre-specified threshold, thereby conserving computational resources and reducing costs.

In addition to uncertainty characterization, the method incorporates a temporal inference mechanism. By integrating historical predictions into the current inference process, we refine the perception model's predictions further, resulting in tighter theoretical bounds and improved accuracy.

We validate the proposed method through extensive quantitative evaluations using the NuScenes driving dataset. The method demonstrates a significant improvement in prediction accuracy by over 10 percent while halving the number of queries made to the foundation model, underscoring the effectiveness of the method in balancing performance and resource efficiency. Moreover, embedding the temporal mechanism in the method further enhances average prediction accuracy by 5 percent under the same number of queries to the foundation model.

\section{Related Work}

Multimodal foundation models \cite{openai2023gpt4, bai2023qwen, llava}, with the ability to process texts and images, have demonstrated impressive capabilities on perception tasks such as object detection \cite{yuan2021florence, zohar2023open, zou2023object}, motion prediction \cite{jiang2024motiongpt, yang2024hard}, image QA and captioning \cite{yu2022coca, yuan2021florence}, and video understanding \cite{video-und-1, video-und-2, video-und-3}. In autonomous driving, recent works \cite{yan2024forging, gao2024survey} show that the foundation models' capabilities on a subset of driving perception tasks have surpassed the state-of-the-art driving perception models, such as UniAD \cite{uniad}, VAD \cite{vad}, ST-P3 \cite{st-p3}, GenAD \cite{zheng2024genad}, etc. However, how to utilize such capabilities to improve driving perception remains challenging.

Existing works \cite{mao2023gpt, liao2024gpt, wen2023road} incorporate multimodal foundation models in autonomous driving, using the reasoning capability of the foundation models to understand visual observation and make high-level driving decisions. Some of these works particularly focus on driving perception \cite{vlm-perception-1, vlm-perception-2, wang2024omnidrive}. They use foundation models as the perception model to interpret the driving scenes and surrounding objects. Furthermore, a few works fine-tune foundation models for self-driving and robot control \cite{finetune-1, finetune-2, finetune-3, finetune-4}. However, these works completely rely on foundation models for perception, they are incapable of dense real-world driving applications due to the high computational and financial cost of these models. In contrast, we propose a method that uses an existing driving perception model, e.g., UniAD, for most perception tasks and selectively triggers the foundation model to refine the perception results. Hence, our method utilizes the foundation model to improve perception without introducing high costs.

\section{Problem Formulation}

Let $M_p: \mathcal{I} \mapsto \mathcal{O} \times [0,1]^3$ be a driving perception model that takes image observations from the image space $\mathcal{I}$ and returns predictions of observed objects from the information space $\mathcal{O}$ and softmax confidence scores between 0 and 1. A confidence score indicates the likelihood of the associated prediction being correct. The information space $\mathcal{O}$ consists of category $Cl = \{\text{car, pedestrian, truck, bicycle, ...}\}$ along with the object's bounding box, attribute $At = \{$vehicle: $\{$moving, stopped, parked$\}$,  pedestrian: $\{$moving, stopped, sitting$\}$, cycle: $\{$without rider, with rider$\} \}$, and tracking $Tr \subset \mathbb{N}$ (natural numbers as object identities, predictions with same tracking id indicating the same object across multiple frames).

Let $M_v: \mathcal{I} \times \mathcal{T} \mapsto \mathcal{T} \times [0,1]$ be a foundation model with vision and language capabilities. The model takes image observations and a prompt in the text space $\mathcal{T}$ as inputs, and returns text-based predictions and confidence scores.

\textbf{Remark: } The driving perception model $M_p$ can run locally in the edge device (e.g., autonomous vehicle) in real-time (response time within 0.5 seconds). The foundation model $M_v$ achieves higher prediction accuracy than $M_p$ (0.87 vs. 0.76 as referred to \cref{tab: compare}) but can not run locally in real-time (response time between 1 to 5 seconds) due to its high computational requirements.

Ideally, we can query $M_v$ for every object prediction to enhance the object prediction accuracy. However, due to the cost of $M_v$, we have a limited frequency of querying it. 
Hence, we need to obtain predictions from the cheaper model $M_p$ and develop a method to use $M_v$ within the frequency limit to enhance the prediction accuracy.

\textbf{Problem: } Given a driving perception model $M_p$ and a foundation model $M_v$, design a method that determines when to trigger $M_v$ and when to override  $M_p$'s prediction by $M_v$'s prediction.

The key to the problem is estimating the uncertainty of the perception model and the foundation model. The method triggers the foundation model only when the perception model $M_p$'s uncertainty is higher than a pre-specified threshold. Furthermore, the method estimates the uncertainty of the foundation model $M_v$ and overrides a prediction only if $M_v$ uncertainty is lower than $M_p$'s uncertainty. The method uses \emph{probabilistic guarantees} to quantitatively characterize models' uncertainties, higher guarantee indicates lower uncertainty. 

\begin{definition}
    \label{def: guarantee}
    Let $\{y_1, ..., y_n\}$ be a set of $n$ predictions returned from a predictive model (either perception model or foundation model), let $\{\overline{y}_1, ..., \overline{y}_n\}$ be the set ground truth labels corresponding to these predictions, a \textbf{probabilistic guarantee} $G \in [0, 1]$ is a theoretical lower bound on the probability of each prediction being correct:
    \begin{equation}
        G \le \lim_{n \rightarrow \infty} \sum_{i=1}^n \textbf{1} (y_i = \overline{y}_i),
    \end{equation}
    where \textbf{1} is the indicator function.
\end{definition}

The probabilistic guarantee serves as an uncertainty metric for assessing the reliability of model predictions in the absence of empirical experiments. This guarantee enables users to make informed decisions regarding which predictions to utilize for downstream tasks. Therefore, for every prediction $y$ returned from either $M_p$ or $M_v$, we will establish a probabilistic guarantee $G$.

\section{Methodology}
We develop a method using uncertainty signals (probabilistic guarantees) and a foundation model to improve an existing driving perception system. Given a pre-trained perception model $M_p$, the method first extracts object predictions with confidence scores from the model. Then, it calibrates each confidence score into a probabilistic guarantee $G_p$. Next, it queries a foundation model $M_v$ only if a guarantee is below a threshold. Finally, it replaces $M_p$'s prediction with $M_v$'s prediction if the guarantee $G_v$ of $M_v$'s prediction is greater than $G_p$. A detailed illustration is in \cref{fig: pipeline}.

\subsection{Confidence Calibration}
\label{sec: calibration}
Neural networks are typically overconfident, hence the confidence score does not indicate the actual probability of the prediction being correct.
Therefore, after obtaining a prediction $y$ with a confidence score $c \in [0, 1]$ from the perception model or the foundation model, the method calibrates $c$ into a probabilistic guarantee $G$. This calibration process relies on the theory of conformal prediction \cite{conformal-prediction}.

\paragraph{Driving Perception Model}
Recall that a driving perception model $M_p$ takes images and returns predictions on object category $Cl$, attribute $At$, and tracking $Tr$. The calibration process of $M_p$ starts from obtaining a \emph{calibration set} $\mathcal{C} = \{X_i, \overline{cl}_i, \overline{at}_i, \overline{tr}_i\}_{i=1}^m$ consisting of $m$ samples. Each sample consists of a set of images $X_i$, the ground truth category $\overline{cl}_i$ of an object, the object's ground truth attribute $\overline{at}_i$, and the ground truth tracking id $\overline{tr}_i$. The calibration set satisfied the following assumption:

\textit{Assumption:} The samples in the calibration set are independently identically distributed (i.i.d.) with the test data.

Then, we build three sets of \emph{nonconformity scores} $N_c, N_a, N_t$ for category, attribute, and tracking separately. A nonconformity score is the confidence score of a wrong prediction.

For each sample in the calibration set, we apply $M_p$ to predict the object categories, attributes, and tracking ids. $M_p$ returns a category prediction $cl_i$ with a confidence $c_i$, an attribute prediction $at_i$ with a confidence $a_i$, and a predicted tracking id $tr_i$ with a confidence $t_i$.

We add $c_i$ into $N_c$ only if $cl_i \neq \overline{cl}_i$ and add $a_i$ into $N_a$ if $at_i \neq \overline{at}_i$. For tracking, let $\overline{tr}_{i-1}$ and $tr_{i-1}$ be the ground truth and predicted tracking ids in the previous time frame, we add $t_i$ to $N_t$ if $(\overline{tr}_{i-1} == \overline{tr}_{i}) \neq (tr_{i-1} == tr_i)$, where ``$==$" returns a boolean value indicating whether the left-hand side is identical to the right-hand side.

After looping over the entire calibration set, we get the three sets of nonconformity scores $N_c, N_a, N_t$. The distribution of nonconformity scores is called \emph{nonconformity distribution}. \Cref{fig: nc-scores} shows the nonconformity distributions of $M_p$.
Next, we estimate the probability density functions of the nonconformity distribution, denoted as $f_c, f_a, f_t$.

In the testing stage, for a set of images $X_{m+1}$ beyond the calibration set, we can get predictions with confidence scores $c_{m+1}$, $a_{m+1}$, and $t_{m+1}$. We calibrate these confidence scores into $G(c_{m+1}, f_c)$, $G(a_{m+1}, f_a)$, and $G(t_{m+1}, f_t)$, where
\begin{equation}
    G(c, f) = \int_0^c f(x) dx.
    \label{eq: calibrate}
\end{equation}
By the theory of conformal prediction, $G(c_{m+1}, f_c)$, $G(a_{m+1}, f_a)$, and $G(t_{m+1}, f_t)$ are the probabilistic guarantees.

\begin{proof}
    The nonconformity distribution is the distribution of confidence scores of wrong predictions. For a given confidence threshold $c^* \in [0, 1]$ and a nonconformity distribution $f$, $\int_{c^*}^1 f(x) dx$ is the proportion of wrong predictions whose confidence score is in $[c^*, 1]$, i.e., $\int_{c^*}^1 f(x) dx = \mathbb{P}($ wrong prediction with confidence $ \in [c^*, 1] ) = \mathbb{P}($correct prediction with confidence $ \in [0, 1-c^*])$.

    Then, $\int_0^{c^*} f(x) dx = 1 - \int_{c^*}^1 f(x) dx = \mathbb{P}($ correct prediction's confidence $\in [1-c^*, 1])$. Denote $C^*$ as a set of predictions with confidence greater than $1-c^*$, so-called a \emph{prediction band}. By the theory of conformal prediction, $\mathbb{P}($correct prediction $ \in C^*) = \mathbb{P}($ correct prediction with confidence $\in [1-c^*, 1])$. 
    
    Let $c^* = 1 - c \text{ if } c > 0.5 \text{ else } c - \lim_{x\rightarrow \infty} \frac{1}{x}$, then $C^*$ is consisting of a single element, which is the prediction with confidence $c$. Then $\mathbb{P}($correct prediction $ \in C^*) = \mathbb{P}($ the prediction with confidence $c$ is correct $) \ge \int_0^{c^*} f(x) dx = G(c, f)$. $G(c, f)$ is the calibrated probabilistic guarantee from the confidence $c$.
    
    According to the assumption, the calibration set is i.i.d. from the test data, hence we can apply the nonconformity distribution from the calibration set to test data. Therefore, $G(c, f) = \int_0^c f(x) dx$ is a probabilistic guarantee for a test data sample whose confidence is $c$.
\end{proof}

\paragraph{Foundation Model}
A foundation model $M_v$ takes a set of images and a textual prompt, returns a textual prediction with a confidence score. For example, the input images look like \cref{fig: example} (the bounding box comes with the category prediction) and the prompt/response is in the following format (input prompts in red and responses in blue):
\begin{lstlisting}[language=completion]
    <prompt>Is the bounding box showing [category or attribute]? Answer Y or N only.</prompt>
    <completion> [Y/N].</completion>
\end{lstlisting}
To calibrate the confidence score, we also need a calibration set $\mathcal{C}' = \{X_i, \overline q_i, \overline r_i\}_{i=1}^m$, where $X_i$ is the image set, $\overline q_i \in \mathcal{T}$ and $\overline r_i \in \mathcal{T}$ are the input prompt and ground truth response (either `Y' or `N'). Then, we construct a set of nonconformity scores $N_v = \{c_i \ |\ r_i \neq \overline{r_i}, \ \ r_i, c_i = M_v(X_i, q_i)\}$.

Next, we can estimate the probability density function $f_v$ of the nonconformity distribution. For every new test data $(X_{m+1}, q_{m+1})$, we calibrate the confidence $c_{m+1}$ to a probabilistic guarantee $G(c_{m+1}, f_v)$ as in \cref{eq: calibrate}.

\subsection{Foundation Model Triggering Mechanism}
\label{sec: trigger}
Let $T$ be a pre-specified uncertainty threshold.
For each prediction, we first calibrate the confidence score from the perception model $M_p$ to a probabilistic guarantee $G_p$. If $G_p \ge T$, we keep this prediction and use it for downstream tasks. If $G_p < T$, we trigger the foundation model $M_v$. In particular, for a given category confidence $c$, if the guarantee $G_p = G(c, f_c) < T$, we query $M_v$ to predict the category. Given an attribute confidence $a$, if $G_p = G(a, f_a) < T$, we query $M_v$ to predict the attribute. The queries will be eventually transformed into binary question-answer format, as in the calibration set.

\paragraph{Prediction Update}
Once we obtain the prediction from $M_v$, we also calibrate its confidence $c_v$ into a guarantee $G_v = G(c_v, f_v)$. If $G_v > G_p$, where $G_p = G(a, f_a) \text{ or } G(c, f_c)$, we replace $M_p$'s category/attribute prediction with $M_v$'s prediction. This step minimizes the possibility that we replace the prediction when $M_v$'s prediction is worse than the original.

\paragraph{Temporal Inference}
Additionally, we propose a temporal inference mechanism to obtain predictions from the perception model and establish stronger probabilistic guarantees. This mechanism takes the predictions of an object across multiple time frames as inputs, and returns an aggregated prediction with a probabilistic guarantee.

\begin{figure}
    \centering
    \includegraphics[width=\linewidth]{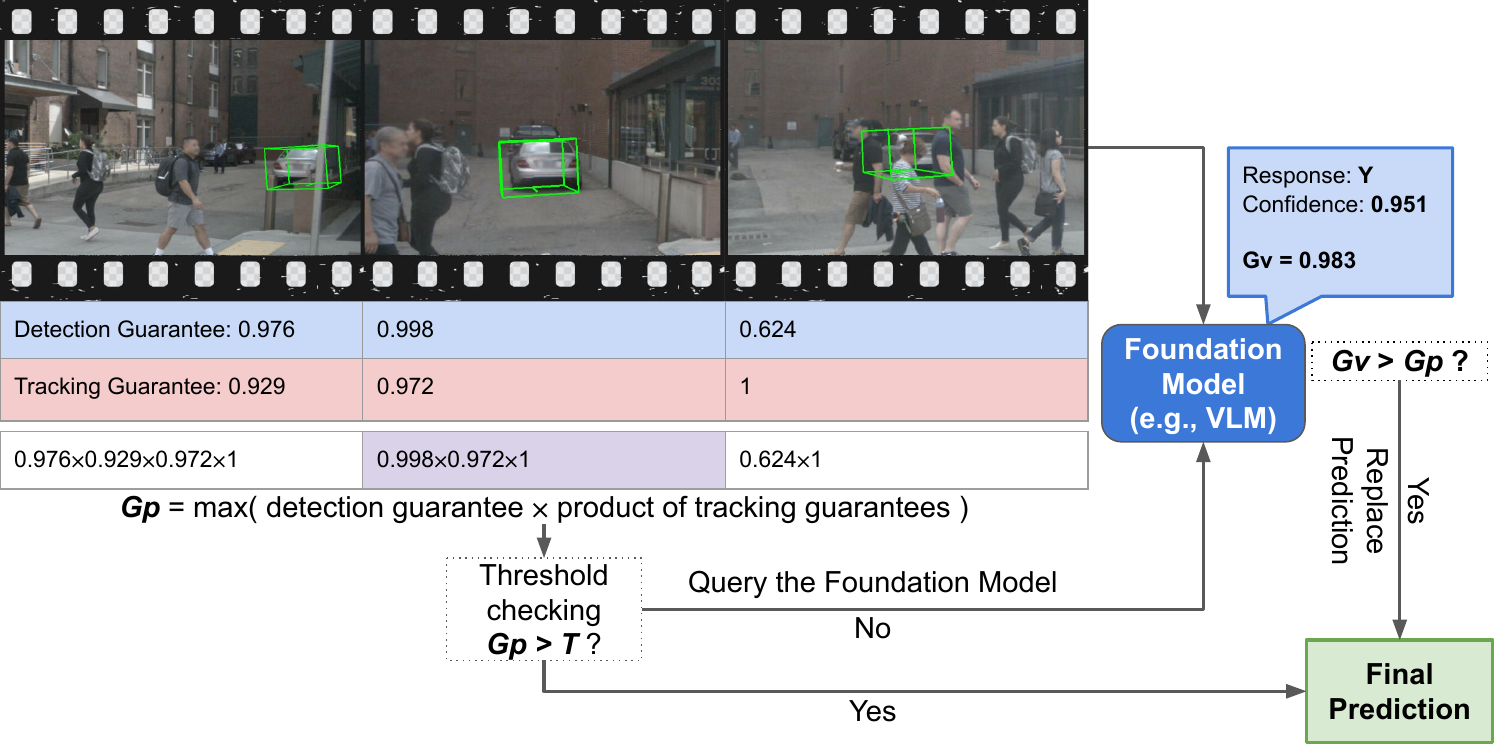}
    \caption{A mechanism that obtains predictions and probabilistic guarantees from objects' temporal information (object tracking and category) across multiple time frames. This mechanism is embedded in Step (2) in \cref{fig: pipeline}.}
    \label{fig: temporal}
    \vspace{-12pt}
\end{figure}

The perception model assigns a tracking id to each predicted object, predictions in different frames with the same tracking id indicating the same object, an example of the same car in different frames is shown in \cref{fig: temporal}. Each tracking id is associated with a confidence score, which we can calibrate it into a probabilistic guarantee that ``the predictions with the same tracking id are indeed corresponding to the same object." We present this calibration $G(t, f_t)$ in \Cref{sec: calibration}.

Let the category and tracking confidence for the current frame be $c_i$ and $t_i$, let the category and tracking confidences for previous $k$ frames be $c_{i-k},..., c_{i-1}$ and $t_{i-k},..., t_{i-1}$. We compute a new probabilistic guarantee $G_p$ of the object category as follows:
\begin{equation}
    \label{eq: temporal-guarantee}
    G_p = \max \{ c_j \times \prod_{l=j}^{i-1} t_{l+1} \}_{j=i-k}^i.
\end{equation}
The aggregated category prediction is the prediction corresponding to the confidence score $c_j$ such that 
\begin{center}
    $G_p = c_j \times \prod_{l=j}^{i-1} t_{l+1}$.
\end{center}

Then, we use the aggregated guarantee $G_p$ instead of the original category guarantee $G(c, t_c)$ and repeat the procedure described in \textit{Prediction Update} in \Cref{sec: trigger}.

We show an illustration of this temporal inference mechanism in \cref{fig: temporal}. In this illustration, $G_p$ is obtained from the prediction in $i-1$'th (middle) frame. 
Empirical evaluations in \cref{fig: vlm-improve} indicate that the temporal inference improves the performance of our method.

\section{Experimental Results}
\label{sec:experiments}

We demonstrate the proposed method and provide quantitative analysis on a state-of-the-art autonomous driving dataset, NuScenes \cite{nuscene}. In the experiments, we use UniAD \cite{uniad} as the driving perception model $M_p$ and GPT-4o-mini \cite{openai2023gpt4} as the multimodal foundation model $M_v$.

\subsection{Nonconformity Distribution}

\begin{figure*}[t]
    \centering
    \includegraphics[width=0.24\linewidth]{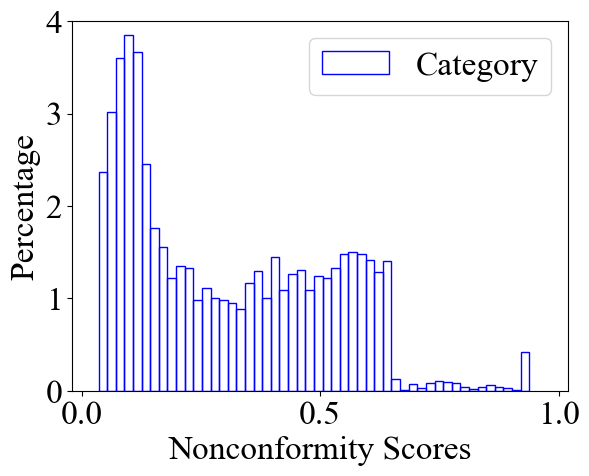}
    \includegraphics[width=0.24\linewidth]{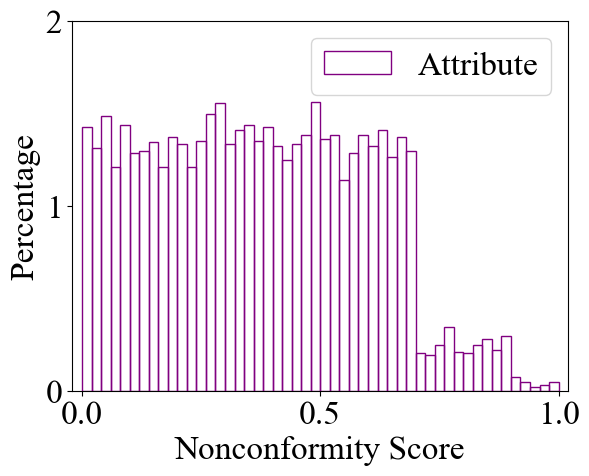}
    \includegraphics[width=0.24\linewidth]{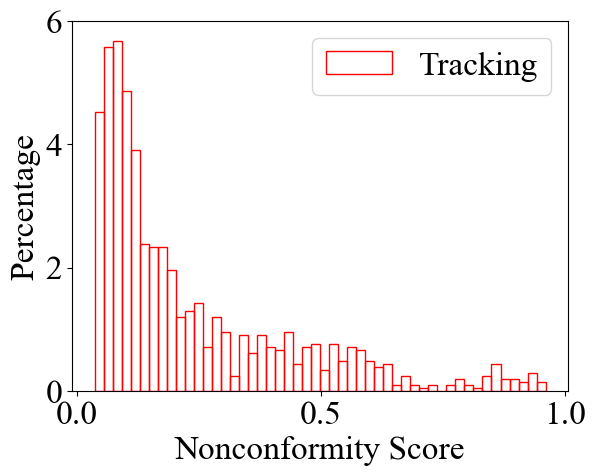}
    \includegraphics[width=0.24\linewidth]{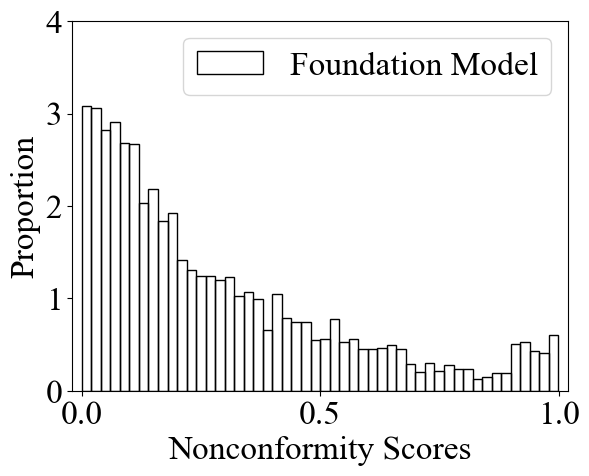}
    \caption{Nonconformity distributions for the UniAD and GPT-4o-mini. We use these distributions from left to right to estimate the probability density functions $f_c, f_a, f_t$ described in \Cref{sec: calibration} and $f_v$ in \Cref{sec: trigger}. Then, we can use \cref{eq: calibrate} or \cref{eq: temporal-guarantee} for confidence calibration.}
    \label{fig: nc-scores}
    \vspace{-10pt}
\end{figure*}

We start by obtaining nonconformity distributions for $M_p$ and $M_v$. For the driving perception model $M_p$, we extract 5215 object predictions from NuScenes as the calibration set. Each prediction includes the object category, attribute, and tracking id. Hence, we follow the procedure described in \Cref{sec: calibration} to obtain three nonconformity distributions for category, attribute, and tracking. 

\Cref{fig: nc-scores} shows the distributions for the perception model. 
Note that nonconformity scores measure the degree to which a data point differs from the existing training data. A high nonconformity score indicates that a point poorly fits the model’s expectations based on the training data, and a low nonconformity score indicates that the point is more typical or expected. Therefore, a right-skewed skewed nonconformity distribution means that most points conform well to the training data, which means the model performs well.

From \cref{fig: nc-scores}, we observe that the category and tracking distributions are right-skewed, and the attribute distribution is close to uniform. Hence, we expect to query the foundation model more frequently and get a large improvement in attribute classification. Moreover, very few points are above 0.7 in all three distributions, meaning the perception model's prediction is reliable if its confidence is above 0.7.


Then, we estimate the three probability density functions $f_c, f_a, f_t$ that will be used for confidence calibration. 
Similarly, we follow the procedure in \Cref{sec: calibration} to obtain a nonconformity distribution for GPT-4o-mini. The calibration set consists of 5215x2 queries regarding object categories and attributes. We present the nonconformity distribution in \cref{fig: nc-scores} and estimate its probability density function $f_v$ for future confidence calibration.

\subsection{Driving Perception Enhancement}
Once we obtain the nonconformity distributions for both models, we follow the procedure described in \Cref{sec: trigger} to improve the perception results. We use the following prompt template when querying the foundation model:
\begin{lstlisting}[language=completion]
    <prompt>[List of images with bounding boxes]</prompt>
    <prompt>What is the bounding box showing? 
    [List of categories or attributes] </prompt>
    <completion> [Choose one category or attribute].</completion>
    ===start a new conversation===
    <prompt>[List of images with bounding boxes]
    Is the bounding box showing [chosen category or attribute]? Answer Y or N only.</prompt>
    <completion> [Y/N].</completion>
\end{lstlisting}
We show an example of the list of images in \cref{fig: temporal}.

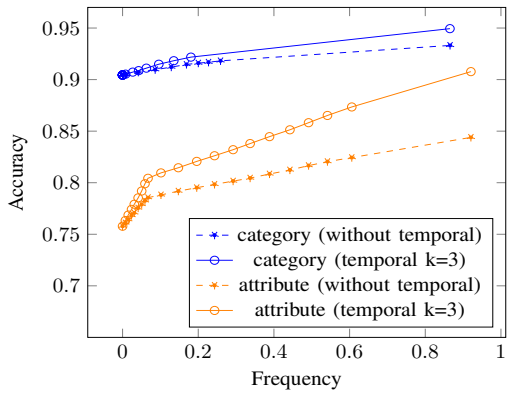
\begin{figure}
    \centering
    \resizebox{0.8\linewidth}{0.6\linewidth}{\begin{tikzpicture}
\begin{axis}[
ylabel=Accuracy,
xlabel=Frequency,
ymin = 0.65,
ymax = 0.97,
ytick = {0.7, 0.75, 0.8, 0.85, 0.9, 0.95},
legend pos=south east,
]
\addplot[dashed, blue, mark=10-pointed star] table [x=Category Count, y=Category Accuracy, col sep=comma] {figures/csv/results-vlm.csv};
\addlegendentry{category (without temporal)}

\addplot[blue, mark=o] table [x=Category Count (Temporal), y=Category Accuracy (Temporal), col sep=comma] {figures/csv/results-vlm.csv};
\addlegendentry{category (temporal k=3)}

\addplot[dashed, orange, mark=10-pointed star] table [x=Attribute Count, y=Attribute Accuracy, col sep=comma] {figures/csv/results-vlm.csv};
\addlegendentry{attribute (without temporal)}

\addplot[orange, mark=o] table [x=Attribute Count (Temporal), y=Attribute Accuracy (Temporal), col sep=comma] {figures/csv/results-vlm.csv};
\addlegendentry{attribute (temporal k=3)}
\end{axis}
\end{tikzpicture}}
    \caption{The frequency of querying the foundation model versus the accuracy of category/attribute predictions. We can improve the attribute accuracy by 10 percent and category accuracy by 3 percent while halving the querying frequency.}
    \label{fig: vlm-improve}
    \vspace{-15pt}
\end{figure}

We evaluate the proposed method over 150 scenes (different from the calibration set) with 82647 object predictions. \Cref{fig: vlm-improve} shows the frequency of querying the foundation model versus the prediction accuracy under different user-specified thresholds. The \emph{accuracy} is the number of correct predictions over the number of all predictions (82647). The \emph{querying frequency} is the number of queries to the foundation model over the number of all predictions (82647).

The method increases the category prediction accuracy from 90 to 93 percent and the attribute accuracy from 75 to 85 percent while keeping the querying frequency lower than 50 percent.

\begin{figure}
    \centering
    \includegraphics[height=0.3\linewidth]{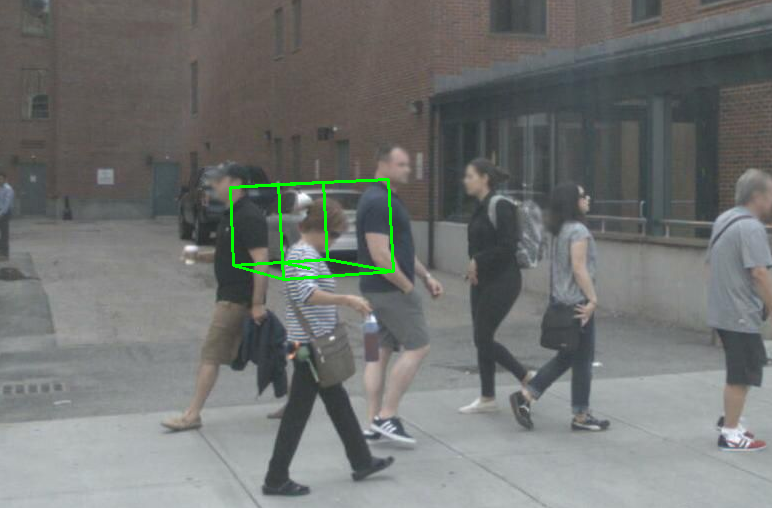}
    \includegraphics[height=0.3\linewidth]{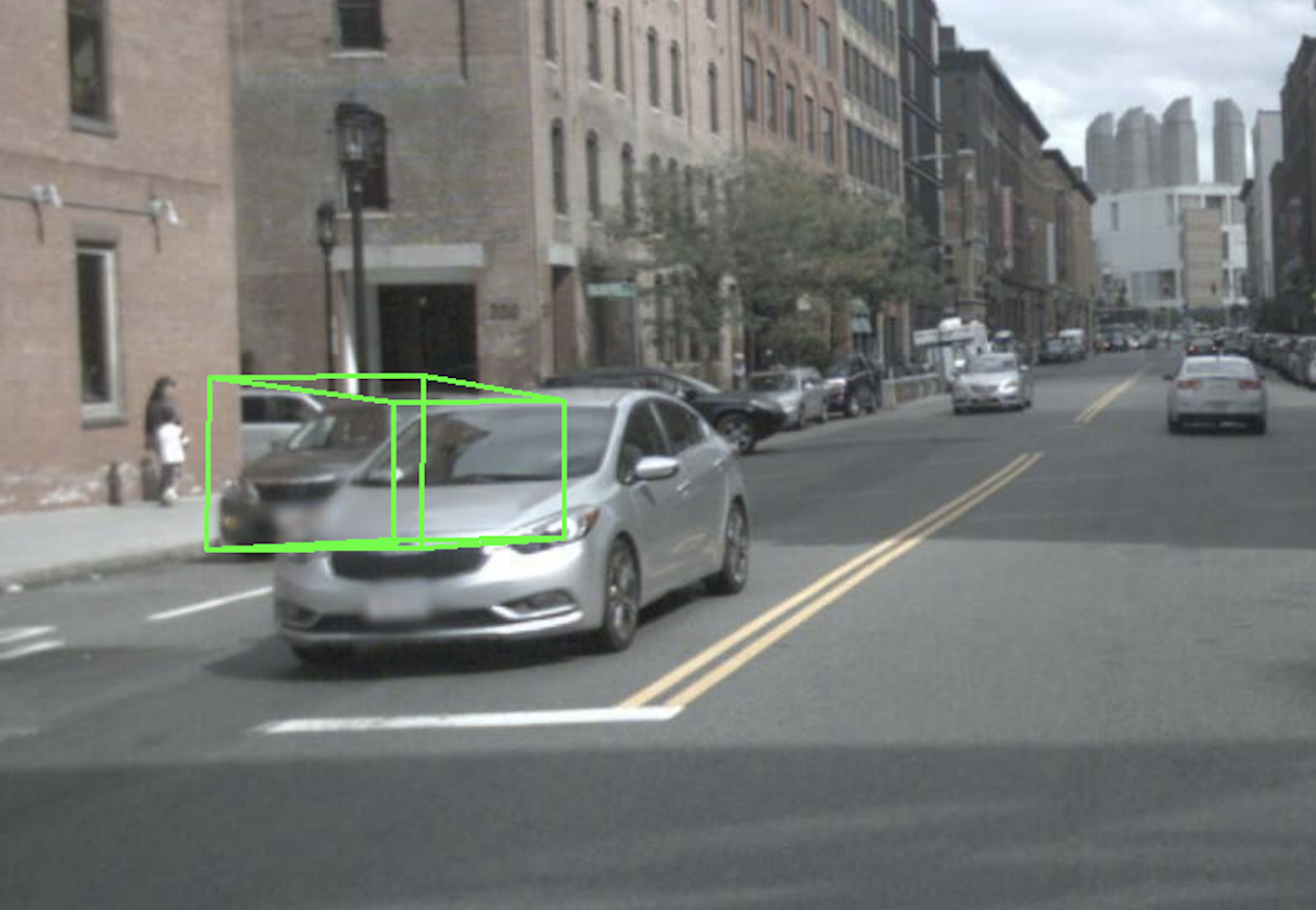}
    \caption{Failure perceptions without temporal inference. Due to occlusions, the left example is misclassified as ``pedestrian" and the right example is misclassified as ``moving."}
    \label{fig: example}
\end{figure}

\paragraph{Temporal Inference}
We explicitly compare our method with and without the temporal inference mechanism. In the method without temporal inference, we calibrate the prediction confidences at the current time frame and use the calibrated confidences to trigger the foundation model. In contrast, the method with temporal inference calibrates confidences through \cref{eq: temporal-guarantee}, which provides a tighter probabilistic guarantee and more accurate prediction. In \cref{fig: vlm-improve}, compared to the method without temporal inference, the temporal inference mechanism (set $k=3$ in \cref{eq: temporal-guarantee}) can improve the attribute accuracy by over 5 percent under the same querying frequency.

\Cref{fig: example} shows examples to demonstrate why the temporal inference mechanism can further enhance prediction accuracy. The figure shows several occluded objects. Without temporal inference, the foundation model predicts category and attribute based on the front objects. As the front objects in the examples are very clear, the foundation model returns predictions with high confidence, leading to unexpected replacements for the original predictions.

\begin{figure*}[t]
    \centering
    \includegraphics[width=0.85\linewidth]{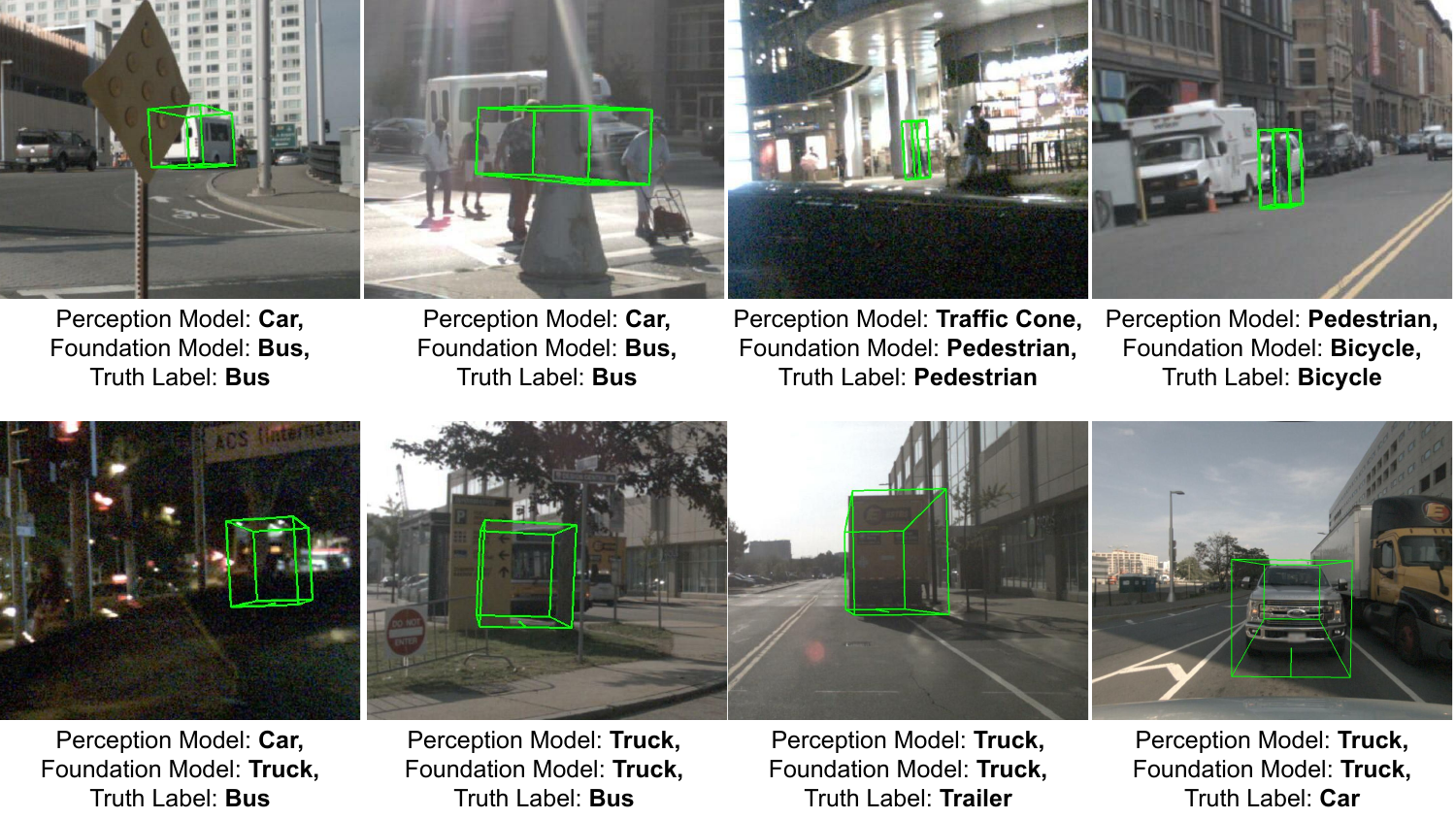}
    \caption{The top row shows examples of prediction errors made by the driving perception model and corrected by the foundation model. The bottom row shows examples of the foundation model failing to refine the predictions. The reasons leading to the foundation model's failures include occlusion, noisy observation, and high similarity between the predicted object and the ground truth object. }
    \label{fig: error-example}
    \vspace{-10pt}
\end{figure*}

We also present some examples on where the perception model makes prediction errors and how the foundation model refines the prediction in \cref{fig: error-example}.

Furthermore, we compare our method with a list of benchmarks in \cref{tab: compare}. Our method achieves the highest attribute accuracy. Although the category accuracy of our method is slightly lower than GPT-4o-mini, we significantly reduce the frequency of querying GPT-4o-mini, hence lowering the computational and financial cost.

\begin{table}
    \centering
    \begin{tabular}{||c|c|c|c|c|c||}
    \hline
          &  Cate & Attr & V Attr & C Attr & P Attr \\
         \hline
         UniAD \cite{uniad} & 0.904 & 0.757 & 0.764& 0.894 & 0.737 \\
         GPT-4o-mini \cite{openai2023gpt4} & \textbf{0.966} & 0.873  & 0.912 & 0.931 & 0.774\\
         Ours (no temp) & 0.933 & 0.844  & 0.914 & 0.935 & 0.769 \\
         \textbf{Ours (temp k=3)} & 0.949 & \textbf{0.908}  & \textbf{0.917} & \textbf{0.941} & 0.782\\
         \textbf{Ours (temp k=9)} & 0.952 & 0.896  & 0.909& 0.928 & \textbf{0.786} \\
    \hline
    \end{tabular}
    \caption{Comparison between the max accuracy ($T=1$) of our method and the benchmarks. The columns indicate prediction accuracies for categories, overall attributes, vehicle, cycle, and pedestrian attributes.}
    \vspace{-15pt}
    \label{tab: compare}
\end{table}

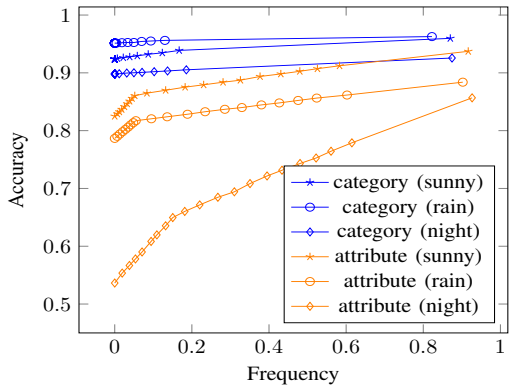
\begin{figure}
    \centering
    \resizebox{0.8\linewidth}{0.6\linewidth}{\begin{tikzpicture}
\begin{axis}[
ylabel=Accuracy,
xlabel=Frequency,
ymin = 0.45,
legend pos=south east,
]
\addplot[blue, mark=star] table [x=Category Count (Sunny), y=Category Accuracy (Sunny), col sep=comma] {figures/csv/condition_results.csv};
\addlegendentry{category (sunny)}

\addplot[blue, mark=o] table [x=Category Count (Rain), y=Category Accuracy (Rain), col sep=comma] {figures/csv/condition_results.csv};
\addlegendentry{category (rain)}

\addplot[blue, mark=diamond] table [x=Category Count (Night), y=Category Accuracy (Night), col sep=comma] {figures/csv/condition_results.csv};
\addlegendentry{category (night)}

\addplot[orange, mark=star] table [x=Attribute Count (Sunny), y=Attribute Accuracy (Sunny), col sep=comma] {figures/csv/condition_results.csv};
\addlegendentry{attribute (sunny)}

\addplot[orange, mark=o] table [x=Attribute Count (Rain), y=Attribute Accuracy (Rain), col sep=comma] {figures/csv/condition_results.csv};
\addlegendentry{attribute (rain)}

\addplot[orange, mark=diamond] table [x=Attribute Count (Night), y=Attribute Accuracy (Night), col sep=comma] {figures/csv/condition_results.csv};
\addlegendentry{attribute (night)}

\end{axis}
\end{tikzpicture}}
    \caption{The frequency of querying the foundation model versus the category/attribute accuracies within scenes under sunny, rainy, and night conditions.}
    \label{fig: condition}
    \vspace{-15pt}
\end{figure}

\paragraph{Scene Analysis}
We select the scenes under sunny, rainy, and night conditions to evaluate the proposed method independently and compare its performance. The experiments in this section use the temporal inference mechanism with $K=3$. \Cref{fig: condition} shows the prediction accuracy under each querying frequency at different conditions. Within the three selected conditions, the perception model initially performs worst at night and the foundation model shows the most significant improvement at night as well. The initial performance and degrees of improvement in sunny and rainy conditions are relatively close. Such results enable users to choose threshold $T$ depending on the weather or lighting conditions, e.g., set a higher $T$ at night.

A notable advantage of the foundation model lies in its strong generalization capability. Our proposed method effectively harnesses this capability, as demonstrated by \cref{fig: condition}, where even a low frequency of querying the foundation model significantly improves the performance across varying driving conditions.

\subsection{Empirical Proof on Probabilistic Guarantee}
\begin{figure}
    \centering
    \resizebox{0.8\linewidth}{0.6\linewidth}{\begin{tikzpicture}
\begin{axis}[
ylabel=Accuracy,
xlabel=Guarantee,
ymin = 0.45,
ymax = 1.0,
xmin = 0.45,
xmax = 0.9,
legend pos=south east,
]
\addplot[scatter, mark=10-pointed star, only marks] table [x=guarantee, y=accuracy, col sep=comma] {figures/csv/guarantee-acc.csv};
\addlegendentry{empirical evaluation}

\addplot[red] table[meta=label] {
x y label
0.45 0.45 a
0.7 0.7 a
0.9 0.9 a
};
\addlegendentry{guarantee = accuracy}

\end{axis}
\end{tikzpicture}}
    \caption{Accuracy versus probabilistic guarantee. The accuracy is above the guarantee consistently, complying with the definition of probabilistic guarantee in \cref{def: guarantee}.}
    \label{fig: guarantee}
    \vspace{-15pt}
\end{figure}
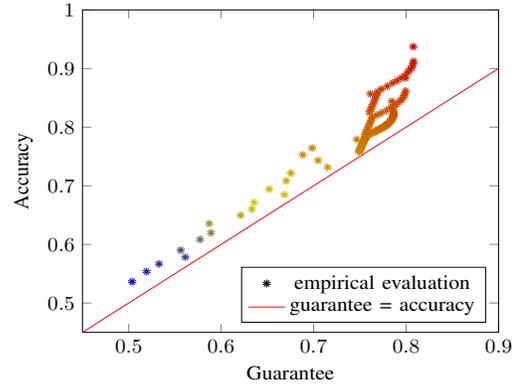

According to \cref{def: guarantee}, the probabilistic guarantee is the lower bound of the accuracy under a large number of samples. We empirically support this claim by computing the accuracy and average probabilistic guarantee under each threshold $T$. Given a set of enhanced predictions, the average guarantee is the average calibrated confidence score and we can evaluate the accuracy. \Cref{fig: guarantee} shows that the accuracy is consistently above the guarantee threshold. Therefore, we claim that under the current number of samples (82647), the probabilistic guarantee holds.

\section{Conclusions}

The proposed method improves driving perception systems by efficiently utilizing multimodal foundation models. By strategically leveraging foundation models only when necessary, based on quantified uncertainties in prediction accuracy, and incorporating a temporal inference mechanism. 
In particular, the method establishes probabilistic guarantees to the predictions, which enable users to determine whether to query the foundation model and use its responses as final predictions.
The method not only enhances the visual prediction accuracy by 10 to 15 percent but also reduces the frequency of resource-intensive model queries by 50 percent.

As a future direction, we will integrate this perception enhancement pipeline with a downstream planning system. We expect an improvement in the plans or trajectories generated from the planning system with respect to the collision rate and the difference between the ground truth trajectory.


\bibliographystyle{IEEEtran}
\bibliography{references}

\begin{thebibliography}{10}
\providecommand{\url}[1]{#1}
\csname url@samestyle\endcsname
\providecommand{\newblock}{\relax}
\providecommand{\bibinfo}[2]{#2}
\providecommand{\BIBentrySTDinterwordspacing}{\spaceskip=0pt\relax}
\providecommand{\BIBentryALTinterwordstretchfactor}{4}
\providecommand{\BIBentryALTinterwordspacing}{\spaceskip=\fontdimen2\font plus
\BIBentryALTinterwordstretchfactor\fontdimen3\font minus \fontdimen4\font\relax}
\providecommand{\BIBforeignlanguage}[2]{{%
\expandafter\ifx\csname l@#1\endcsname\relax
\typeout{** WARNING: IEEEtran.bst: No hyphenation pattern has been}%
\typeout{** loaded for the language `#1'. Using the pattern for}%
\typeout{** the default language instead.}%
\else
\language=\csname l@#1\endcsname
\fi
#2}}
\providecommand{\BIBdecl}{\relax}
\BIBdecl

\bibitem{openai2023gpt4}
OpenAI, ``Gpt-4 technical report,'' \emph{arXiv preprint arXiv:2303.08774}, 2023.

\bibitem{bai2023qwen}
J.~Bai, S.~Bai, Y.~Chu, Z.~Cui, K.~Dang, X.~Deng, Y.~Fan, W.~Ge, Y.~Han, F.~Huang \emph{et~al.}, ``Qwen technical report,'' \emph{arXiv preprint arXiv:2309.16609}, 2023.

\bibitem{llava}
H.~Liu, C.~Li, Q.~Wu, and Y.~J. Lee, ``Visual instruction tuning,'' in \emph{Advances in Neural Information Processing Systems}, A.~Oh, T.~Naumann, A.~Globerson, K.~Saenko, M.~Hardt, and S.~Levine, Eds., New Orleans, LA, USA, 2023.

\bibitem{yuan2021florence}
L.~Yuan, D.~Chen, Y.-L. Chen, N.~Codella, X.~Dai, J.~Gao, H.~Hu, X.~Huang, B.~Li, C.~Li \emph{et~al.}, ``Florence: A new foundation model for computer vision,'' 2021.

\bibitem{zohar2023open}
O.~Zohar, A.~Lozano, S.~Goel, S.~Yeung, and K.-C. Wang, ``Open world object detection in the era of foundation models,'' 2023.

\bibitem{zou2023object}
Z.~Zou, K.~Chen, Z.~Shi, Y.~Guo, and J.~Ye, ``Object detection in 20 years: A survey,'' \emph{Proceedings of the IEEE}, vol. 111, no.~3, pp. 257--276, 2023.

\bibitem{jiang2024motiongpt}
B.~Jiang, X.~Chen, W.~Liu, J.~Yu, G.~Yu, and T.~Chen, ``Motiongpt: Human motion as a foreign language,'' \emph{Advances in Neural Information Processing Systems}, vol.~36, 2024.

\bibitem{yang2024hard}
Y.~Yang, Q.~Zhang, K.~Ikemura, N.~Batool, and J.~Folkesson, ``Hard cases detection in motion prediction by vision-language foundation models,'' 2024.

\bibitem{yu2022coca}
J.~Yu, Z.~Wang, V.~Vasudevan, L.~Yeung, M.~Seyedhosseini, and Y.~Wu, ``Coca: Contrastive captioners are image-text foundation models,'' \emph{arXiv preprint arXiv:2205.01917}, 2022.

\bibitem{video-und-1}
S.~Buch, C.~Eyzaguirre, A.~Gaidon, J.~Wu, L.~Fei{-}Fei, and J.~C. Niebles, ``Revisiting the "video" in video-language understanding,'' in \emph{{IEEE/CVF} Conference on Computer Vision and Pattern Recognition}.\hskip 1em plus 0.5em minus 0.4em\relax New Orleans, LA, USA: {IEEE}, 2022, pp. 2907--2917.

\bibitem{video-und-2}
C.~Ju, T.~Han, K.~Zheng, Y.~Zhang, and W.~Xie, ``Prompting visual-language models for efficient video understanding,'' in \emph{Computer Vision - {ECCV} 2022 - 17th European Conference}, ser. Lecture Notes in Computer Science, S.~Avidan, G.~J. Brostow, M.~Ciss{\'{e}}, G.~M. Farinella, and T.~Hassner, Eds., vol. 13695.\hskip 1em plus 0.5em minus 0.4em\relax Tel Aviv, Israel: Springer, 2022, pp. 105--124.

\bibitem{video-und-3}
H.~Xu, G.~Ghosh, P.-Y. Huang, P.~Arora, M.~Aminzadeh, C.~Feichtenhofer, F.~Metze, and L.~Zettlemoyer, ``Vlm: Task-agnostic video-language model pre-training for video understanding,'' 2021.

\bibitem{yan2024forging}
X.~Yan, H.~Zhang, Y.~Cai, J.~Guo, W.~Qiu, B.~Gao, K.~Zhou, Y.~Zhao, H.~Jin, J.~Gao \emph{et~al.}, ``Forging vision foundation models for autonomous driving: Challenges, methodologies, and opportunities,'' 2024.

\bibitem{gao2024survey}
H.~Gao, Y.~Li, K.~Long, M.~Yang, and Y.~Shen, ``A survey for foundation models in autonomous driving,'' 2024.

\bibitem{uniad}
Y.~Hu, J.~Yang, L.~Chen, K.~Li, C.~Sima, X.~Zhu, S.~Chai, S.~Du, T.~Lin, W.~Wang, L.~Lu, X.~Jia, Q.~Liu, J.~Dai, Y.~Qiao, and H.~Li, ``Planning-oriented autonomous driving,'' in \emph{{IEEE/CVF} Conference on Computer Vision and Pattern Recognition}.\hskip 1em plus 0.5em minus 0.4em\relax Vancouver, BC, Canada: {IEEE}, 2023, pp. 17\,853--17\,862.

\bibitem{vad}
B.~Jiang, S.~Chen, Q.~Xu, B.~Liao, J.~Chen, H.~Zhou, Q.~Zhang, W.~Liu, C.~Huang, and X.~Wang, ``{VAD:} vectorized scene representation for efficient autonomous driving,'' in \emph{{IEEE/CVF} International Conference on Computer Vision}.\hskip 1em plus 0.5em minus 0.4em\relax Paris, France: {IEEE}, 2023, pp. 8306--8316.

\bibitem{st-p3}
S.~Hu, L.~Chen, P.~Wu, H.~Li, J.~Yan, and D.~Tao, ``{ST-P3:} end-to-end vision-based autonomous driving via spatial-temporal feature learning,'' in \emph{Computer Vision - {ECCV} 2022 - 17th European Conference}, ser. Lecture Notes in Computer Science, S.~Avidan, G.~J. Brostow, M.~Ciss{\'{e}}, G.~M. Farinella, and T.~Hassner, Eds., vol. 13698.\hskip 1em plus 0.5em minus 0.4em\relax Tel Aviv, Israel: Springer, 2022, pp. 533--549.

\bibitem{zheng2024genad}
W.~Zheng, R.~Song, X.~Guo, and L.~Chen, ``Genad: Generative end-to-end autonomous driving,'' \emph{arXiv preprint arXiv:2402.11502}, 2024.

\bibitem{mao2023gpt}
J.~Mao, Y.~Qian, H.~Zhao, and Y.~Wang, ``Gpt-driver: Learning to drive with gpt,'' \emph{arXiv preprint arXiv:2310.01415}, 2023.

\bibitem{liao2024gpt}
H.~Liao, H.~Shen, Z.~Li, C.~Wang, G.~Li, Y.~Bie, and C.~Xu, ``Gpt-4 enhanced multimodal grounding for autonomous driving: Leveraging cross-modal attention with large language models,'' \emph{Communications in Transportation Research}, vol.~4, p. 100116, 2024.

\bibitem{wen2023road}
L.~Wen, X.~Yang, D.~Fu, X.~Wang, P.~Cai, X.~Li, T.~Ma, Y.~Li, L.~Xu, D.~Shang \emph{et~al.}, ``On the road with gpt-4v (ision): Early explorations of visual-language model on autonomous driving,'' \emph{arXiv preprint arXiv:2311.05332}, 2023.

\bibitem{vlm-perception-1}
M.~Najibi, J.~Ji, Y.~Zhou, C.~R. Qi, X.~Yan, S.~Ettinger, and D.~Anguelov, ``Unsupervised 3d perception with 2d vision-language distillation for autonomous driving,'' in \emph{{IEEE/CVF} International Conference on Computer Vision}.\hskip 1em plus 0.5em minus 0.4em\relax Paris, France: {IEEE}, 2023, pp. 8568--8578.

\bibitem{vlm-perception-2}
S.~Jain, S.~Thapa, K.~Chen, A.~L. Abbott, and A.~Sarkar, ``Semantic understanding of traffic scenes with large vision language models,'' in \emph{{IEEE} Intelligent Vehicles Symposium}.\hskip 1em plus 0.5em minus 0.4em\relax Jeju Island, Republic of Korea: {IEEE}, 2024, pp. 1580--1587.

\bibitem{wang2024omnidrive}
S.~Wang, Z.~Yu, X.~Jiang, S.~Lan, M.~Shi, N.~Chang, J.~Kautz, Y.~Li, and J.~M. Alvarez, ``Omnidrive: A holistic llm-agent framework for autonomous driving with 3d perception, reasoning and planning,'' 2024.

\bibitem{finetune-1}
Y.~Yang, N.~P. Bhatt, T.~Ingebrand, W.~Ward, S.~Carr, A.~Wang, and U.~Topcu, ``Fine-tuning language models using formal methods feedback: {A} use case in autonomous systems,'' in \emph{Proceedings of the Seventh Annual Conference on Machine Learning and Systems}.\hskip 1em plus 0.5em minus 0.4em\relax Santa Clara, CA, USA: mlsys.org, 2024.

\bibitem{finetune-2}
\BIBentryALTinterwordspacing
Y.~Wang, Z.~Huang, Q.~Liu, Y.~Zheng, J.~Hong, J.~Chen, L.~Xiong, B.~Gao, and H.~Chen, ``Drive as veteran: Fine-tuning of an onboard large language model for highway autonomous driving,'' in \emph{{IEEE} Intelligent Vehicles Symposium}.\hskip 1em plus 0.5em minus 0.4em\relax Jeju Island, Republic of Korea: {IEEE}, 2024, pp. 502--508. [Online]. Available: \url{https://doi.org/10.1109/IV55156.2024.10588851}
\BIBentrySTDinterwordspacing

\bibitem{finetune-3}
D.~Hu, C.~Huang, J.~Wu, and H.~Gao, ``Pre-trained transformer-enabled strategies with human-guided fine-tuning for end-to-end navigation of autonomous vehicles,'' \emph{arXiv preprint arXiv:2402.12666}, 2024.

\bibitem{finetune-4}
X.~Li, M.~Liu, H.~Zhang, C.~Yu, J.~Xu, H.~Wu, C.~Cheang, Y.~Jing, W.~Zhang, H.~Liu \emph{et~al.}, ``Vision-language foundation models as effective robot imitators,'' \emph{arXiv preprint arXiv:2311.01378}, 2023.

\bibitem{conformal-prediction}
G.~Shafer and V.~Vovk, ``A tutorial on conformal prediction.'' \emph{Journal of Machine Learning Research}, vol.~9, no.~3, 2008.

\bibitem{nuscene}
H.~Caesar, V.~Bankiti, A.~H. Lang, S.~Vora, V.~E. Liong, Q.~Xu, A.~Krishnan, Y.~Pan, G.~Baldan, and O.~Beijbom, ``nuscenes: {A} multimodal dataset for autonomous driving,'' in \emph{{IEEE/CVF} Conference on Computer Vision and Pattern Recognition}.\hskip 1em plus 0.5em minus 0.4em\relax Seattle, WA, USA: Computer Vision Foundation / {IEEE}, 2020, pp. 11\,618--11\,628.

\end{thebibliography}

\end{document}